# Physical Computing With No Clock to Implement the Gaussian Pyramid of SIFT Algorithm

Yi Li, Qi Wei, Fei Qiao, Huazhong Yang
Tsinghua University, Beijing, China
E-mail: qiaofei@tsinghua.edu.cn

**Abstract** - Physical computing is a technology utilizing the nature of electronic devices and circuit topology to cope with computing tasks. In this paper, we propose an active circuit network to implement multi-scale Gaussian filter, which is also called *Gaussian Pyramid* in image preprocessing. Various kinds of methods have been tried to accelerate the key stage in image feature extracting algorithm these years. Compared with existing technologies, GPU parallel computing and FPGA accelerating technology, physical computing has great advantage on processing speed as well as power consumption. We have verified that processing time to implement the Gaussian pyramid of the SIFT algorithm stands on nanosecond level through the physical computing technology, while other existing methods all need at least hundreds of millisecond. With an estimate on the stray capacitance of the circuit, the power consumption is around 670pJ to filter a 256x256 image. To the best of our knowledge, this is the most fast processing technology to accelerate the SIFT algorithm, and it is also a rather energy-efficient method, thanks to the proposed physical computing technology.

## I Introduction

After Analog-to-information converter (AIC) was proposed in 2004, the new technology obtained a boom growth in the following years. Many systems, especially those operating in the radio frequency (RF) bands, severely stress conventional analog-to-digital converter (ADC) technologies [1]. In most applications, signal is firstly converted and stored as digital data and then processed to extract important features. Unfortunately, this would do lots waste in terms of power consumption and processing time. Directly coping with the signal at the source is a new directory to replace current ADCs in some applications, which could further save power and processing requirements [2], [1].

Emerging imaging applications present new challenges on underlying computing technologies. State-of-the-art digital-based technologies, including software-platform algorithm optimization and some FPGA accelerating technologies in hardware-platform, cannot cover the requirements of embedded systems in some aspects. We have investigated many efforts aiming at accelerating image feature extracting algorithms. Scale-invariant feature transform (SIFT) is a classical algorithm in computer vision to detect and describe local features in images. SIFT algorithm was proposed by Lowe in 1999, which has been considered as one of the most robust approaches among all the feature detectors [4]. The SIFT features are invariant to:1) scale; 2) rotation; and 3)illumination, that are essential in many applications such as object detection and robot navigation. Thanks to the high computational cost of multi-scale convolution, it is rather difficult to resolve SIFT algorithm in real time, that prevents the SIFT being utilized in robot navigation and many other embedded systems [5].

TABLE I
RUN TIME ANALYSIS OF A SIFT SOFTWARE PROGRAM FOR A VGA "BEAVER" IMAGE ON INTEL CORE 2 DUO2.09 GHZ CPU [5]

| Steps in Algorithm | Run Time(s) | Percentage |
|---|---|---|
| Gaussian pyramid construction | 2.1 | 72.85% |
| DoG space construction | 0.004 | 0.14% |
| Key-points detection | 0.16 | 5.71% |
| Gradient and orientation assignment | 0.08 | 2.85% |
| Descriptors generation | 0.52 | 18.57% |
| Entire SIFT operation | 2.87 | 100% |

Lots of speed-up feature detectors have been proposed based on software platform like SURF[6] and fast SIFT[7]. They try to find a new balance between the precision and computing speed. Sacrificing the quality of features extracted is a general way for the algorithms to reduce the computational time. Even so, the processing speed may not meet the real-time requirement in many embedded systems. Another technology was proposed to specially deal with digital image processing tasks—Graphic processing unit (GPU). This is a typical method to accelerate image processing through parallel computing. Except all the methods introduced above based on software platform, some efforts have been tried on hardware platform. Various tries have been made on filedprogrammable gate array(FPGA) boards according to SIFT algorithm characteristics. Generally, we can roughly divide SIFT into two main modules, feature detection part including Gaussian Pyramid establishing and key points descriptor generation part. Considering that the Gaussian Pyramid establishing takes around 80% of the whole algorithm computational time, previous researchers realized the feature detection part with hardware to gain a obvious speed-up, whereas the feature descriptor generation is done by software. A work from Taiwan university in 2011 realized SIFT with hardware including both parts of it, that accelerates this algorithm more than previous approches. Table I shows the run time analysis of executing an SIFT software program on a 2.1GHz Intel CPU. It takes around 2.87s to run a VGA image with 640x480 pixels. SIFT algorithm can be divided into five detailed parts, among which previous three parts belong to feature detection stage

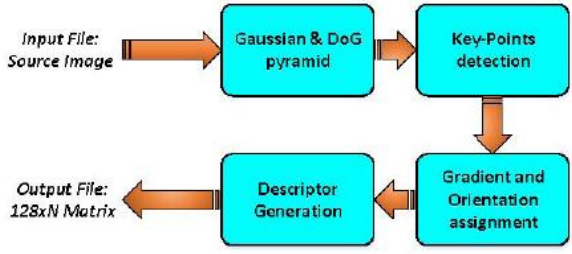

Fig. 1. SIFT algorithm flow consists of four components: Gaussian and DoG pyramid, Key-points detection, gradient and orientation assignment, descriptor generation

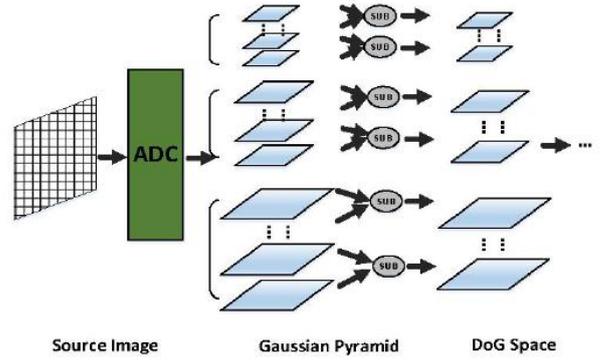

Fig. 2. Gaussian Pyramid and DoG space: Analog source signals are converted to digital bits through ADC, and subsequent processing is in the digital domain

and the left belong to featrure descriptors generation stage. The first step of SIFT operation, Gaussian pyramid construction, costs 72.85% computational time of the whole algorithm [5].

In this paper, based on the background of physical computing, we propose a circuit architecture to cope with the key component of SIFT algorithm. Different from conventional digital processing methods ranging from software algorithm like SURF to hardware techology such as FPGA, physical computing gets rid of the limit of clock absolutely. Also, unlike analog computer, there is no complex connection between resistors, capacitances and transistors to build analog Arithmetic Logical Unit(ALU). Physical computing focuses on the characteristics of the device itself and the topology of circuit architecture. Analog signal is dealt with at the source-end before transferred to digital signal. We utilize an active resistor network[8] to implement Gaussian filtering to the voltage signal obtained from camera. Every pixel corresponding to the camera maps to the nodes of the circuit network, which plays the role of the source to stimulate the whole system. Only nanosecond-level time need to prepare Output-voltage on every node. In section II the SIFT algorithm will be presented. We will detail our design in section III and section IV contains experiments and the results. Conclusion is given in section V.

## II. SIFT Algorithm Introduction

This section intentionally to describe the SIFT algorithm. Generally, the input is a digital image. After the SIFT processing, a number of descriptors are produced, each of which is a 128 dimension vector. The output data is a matrix whose size is 128xN (N refers to the number of features extracted). The algorithm can be divided into four basic stages as the Figure 1 shows: (a) to establish the multi-scale Gaussian and Difference of Gaussian (DoG) pyramid; (b) to detect and refine those robust key-points; (c) to compute and estimate the principle orientation of each key-point; (d) to generate the feature descriptors and store them as a 128xN matrix.

### A. Gaussian Pyramid And DoG Space Construction

This stage is a rather critical component and is also the first step of the SIFT algorithm. The source image is I(x, y), and the

Gaussian kernel function is denoted as G(x, y, σ ), where the x and y are horizontal and vertical coordinates in physical space respectively, and σ can be treated as the coordinate in scale-space. The output data is a Gaussian-convolved image, denoted as M(x, y, σ ). The best results will be obtained if your computer word-processor has several font sizes. Try to follow the font sizes specified in Table I as best as you can. As an aid to gauging font size, 1 point is about 0.35mm. Use a proportional, serif font such as Times of Dutch Roman.

$$M(x,y,\sigma) = G(x,y,\sigma) * I(x,y) \quad (1)$$

Where G(x, y , σ) is the Gaussian kernel function

$$G(x,y,\sigma) = \frac{1}{2\pi\sigma^2}e^{-(x^2+y^2)} \quad (2)$$

A DoG image is the difference of contiguous filtered images, illustrated in figure 2. For instance, the difference of two images filtered by σ and kσ is

$$D(x,y,\sigma) = M(x,y,k\sigma) - M(x,y,\sigma) \quad (3)$$

According to Lowe's derivation, the parameter $k=2^{\frac{1}{S}}$, where S equals to the number of filtered images in every octave.

As shown in figure 2, each octave includes S (in Lowe's design S equals to 6) filtered images. Upper-level octaves are down-sampled images of the lower octaves. The number of octave is various according to the size of image. It should make sure that the size of the image on the top of the pyramid is at least 8 _ 8. In every octave, the images are filtered by Gaussian kernel functions, the scale of which is ranging from σ to kσ. That is, the filtered images in each octave can be denoted as {G(σ), G(kσ), G($k^2$σ), G($k^3$σ), G($k^4$σ), G($k^5$σ)}. Assuming source image is 256 _ 256, the bottom-images of the pyramid is the original size, the upper octave images are down sampled from the first level octave to 128 _ 128. In our design, we just build three octaves, that is, the top octave images of the pyramid are 64x64.

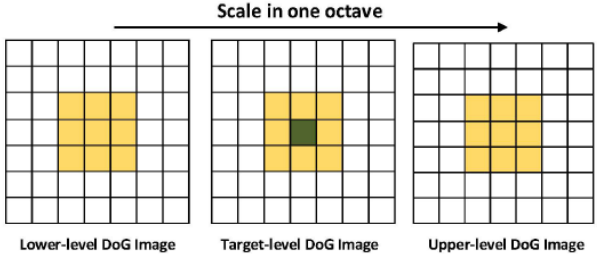

Fig. 3. Key-point detection: The center pixel is compared with other 26 pixels, including its eight surrounding neighbors in the same DoG image, and the nine surrounding neighbors in its lower-level DoG image and nine surrounding neighbors in its upper-level DoG image.

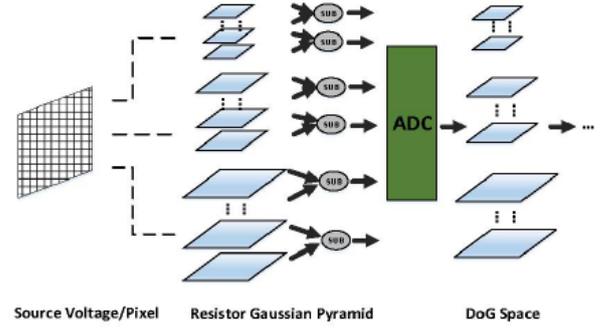

Fig. 4. Analog Gaussian Filter used in SIFT Algorithm: anlog signals are sampled from source-end, and then directly flow through the resistor network the filtered data is the filtered result. The following key-points detection and descriptors generation are processed in digital domain.

The DoG space is obtained from Gaussian pyramid. For instance, the bottom image in DoG space is the difference result of $G(k\sigma)$ and $G(\sigma)$. Similarly, the upper ones are also computed through the difference of two neighbor images. In DoG space, there are $S-1$ differential images in one octave, where S is the number of images in one octave in Gaussian pyramid.

*B. Key-Points Detection*

The key-points are detected from DoG images by finding the local maximum or minimum gray value. Every pixel in one DoG image is compared with other 26 pixels, including its eight surrounding neighbors in the same DoG image, and the nine surrounding neighbors in its lower-level DoG and the nine surrounding neighbors in its upper-level DoG. Once the pixel is the maxima or minima among the 27 pixels, it is identified as a key-point.

*C. Gradient and Orientation assignment*

Every key-point detected should have an orientation, which is calculated by the neighbors in the same Gaussian-filtered image. Key-points are found in the DoG images, whereas the orientation computing task is done in the Gaussian filtered images. The orientation of a key-point is relative to its neighbors within the key-radius. An orientation histogram is taken to compute the key-point orientation. Each pixel has its horizontal and vertical gradient, through which the orientation of this pixel will be obtained.

*D. Descriptor generation*

This stage is to generate a vector to describe the key-points and their neighbors. In general, 4x4 pixels are taken to describe the features of the centering point—key point. With each pixel has eight orientation possibilities, the length of the vector is denoted as $16 \times 8 = 128$. So if N key-points are extracted, the SIFT algorithm will output a 128xN matrix.

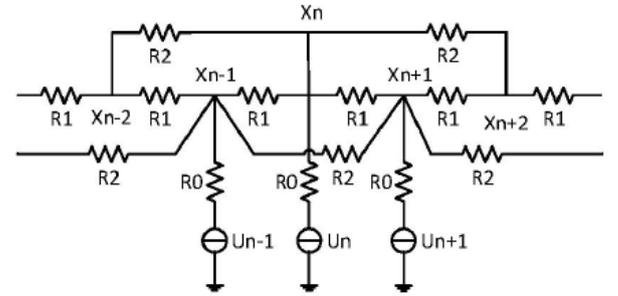

Fig. 5. 1-D Gaussian resistor circuit: $R_2$s are negative resistors, $R_2/R_1 = -4$, $R_0/R_1 = \lambda$. $U_n, U_{n-1}, U_{n+1}$ are input voltages and $x_n, x_{n-1}, x_{n+1}$ are output data.[9]

### III. Active Resistor Gaussian Filter Design

This section we will detail the design of active resistor Gaussian filter, including the principle of the circuit network, and the framework of resistor Gaussian pyramid. As figure 4 illustrating, the conventional Gaussian pyramid stage was replaced with active resistor network. The input data is no longer digital bit, source voltage signals from sensors are directly sampled into the analog circuit instead. After the circuit tends to be stable, node voltages are extracted as the output data.

*A. Active Resistor Network Introduction*

For the resistor network as a Gaussian filter, a strict derivation has been concluded in [8]. A 1-D Gaussian filter can be obtained from 5, where the node $x_n$ holds the output voltage, and the voltage $u_{n-1}$, $u_n$ and $u_{n+1}$ are the input voltages. A equation about the relation between input and output can be obtained according to Kirchhoff current and voltage laws

$$u_n = \frac{1}{g_0}\{(g_0+2g_1+2g_2)x_n - g_1(x_{n-1}+x_{n+1}) - g_2(x_{n-2}+x_{n+2})\} \quad (4)$$

where the g0, g1 and g2 are the conductances in the circuit network, whose values are equal to the reciprocal of their corresponding resistors' values. Abidi [8] and his team utilized a perfect mathematical formula verified by Poggio et al. [10] to a circuit system. Similar with the mathematical theory, minimizing an energy function E in the circuit, which is defined as the mean square difference between the interpolating function (corresponding to the output node voltages) and the samples(corresponding to the input voltages), can obtain a fitting function of the input data. This function can be denoted as

$$E = \sum_i (U(x=i) - V_i)^2 + \lambda \int (\frac{d^2U}{dx^2})^2, dx \quad (5)$$

where the second order partial derivative is a critical faction, and the $\lambda$ is the penalty to control the fluctuation of U(x). Poggio [10] has verified that U(x) is the result of convolution from V and R(x,λ), where R(x,λ) is

$$R(x, \lambda) = \frac{1}{2\lambda^{1/4}} e^{|x|/\sqrt{2}\lambda^{1/4}} cos(\frac{|x|}{\sqrt{2}\lambda^{1/4}} - \frac{\pi}{4}) \quad (6)$$

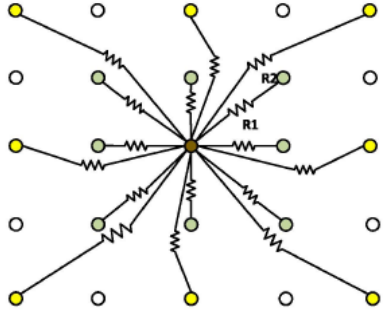

Fig. 6. 2-D Gaussian resistor circuit: to be circle-symmetry, it is extended from 1-D network not only on horizontal and verticle directions, but also on diagonal directions. [9]

The equation R(x, λ) can be approximately treated as a Gaussian kernel function, the parameter λ of which is roughly proportional to the Gaussian width $\sigma^4$. To define the relationship between g0, g1 and g2 in equation 4, equation 5 need to do some transformation,

$$E = \sum_i (U_i - V_i)^2 + \lambda \sum_i (U_{i+1} + U_{i-1} - 2U_i)^2 \quad (7)$$

where i is the index in discrete domain. For equation 7, we can obtain the minima of it by calculate the partial derivative for variable $U_i$. E can get the minima when $\frac{\dot{E}}{U_i} = 0$.

$$0 = 2(U_i - V_i) + \lambda \frac{\partial}{\partial U_i} \sum_k (U_{k-1} + U_{k+1} - 2U_k)^2 \quad (8)$$

$$0 = (U_i - V_i) + \lambda(6U_i - 4(U_{i-1} + U_{i+1}) + (U_{i-2} + U_{i+2})) \quad (9)$$

Equation 9 describes the relationship between U and V, which are the output and input of the resistor network respectively. According to the conclusion in [8], once U and V satisfy the relationship as equation 9, U can be considered as the result of the convolution from V and a Gaussian kernal function. Comparing equation 9 and 4, the proportional relationship between g0, g1 and g2 can be easily calculated,

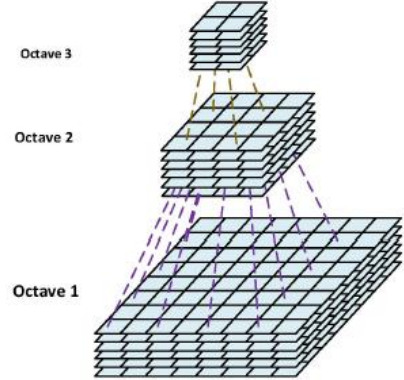

Fig. 7. 2-D Gaussian pyramid circuit: The basic component of the pyramid is 2-D active resistor network. There are three octaves in the pyramid, and six 2-D resistor networks with different $r_0$ make up a octave. Every node in the upper-level octave network connects the nodes between every two pixels in lower-level octave.

$$\frac{g_1}{g_0} = \lambda, \frac{g_1}{g_2} = -4 \quad (10)$$

Similar with 1-D resistor network, the extension to 2-D is not that complex. However, the extension does not just do the same work as 1-D network on both horizontal and vertical direction, moreover, the circle symmetry should be adequately assured in 2-D network. Besides horizontal and vertical directions, the connection relation in figure 6 considers the diagonals as well.

*B. Analog Circuit Gaussian Pyramid*

Sensors transform illumination into voltage signals at the input-end, where every pixel corresponds to a voltage source. For the images in bottom octave of the Gaussian pyramid, each pixel connects to the node in 2-D resistor network. The second octave connects the voltage source between every two pixels along both horizontal and vertical directions, thus implementing the goal of down-sample. The third octave networks connects the voltage source between every four pixels to implement four-fold down sample. The circuit design is showed in figure 7. In our design, there are six images in every octave, every image in which is filtered by different Gaussian filter. We permanently assign the number of octave as three, and experiments have verified that three octave can satisfy most applications' needs.

IV. Experiments

The active resistor circuit network is designed on the background of analog circuit natural advantage. Different from digital circuit, which needs a clock counter to measure the running speed of the system. The analog circuit just spends nanoseconds to reach the stable state, that is also the computing time of this analog circuit spends. The circuit is designed to simulate Gaussian kernel function. Considering the noise in analog circuit and the topology of this network, we design some experiments to analyze the deviation between the analog circuit and ideal Gaussian kernel function. Also, we will also analyze the computing speed and power consumption of the physical architecture.

*A. 1-D Gaussian Resistor Network Verification*

We have analyzed the relation between the parameter $\lambda$ and $\sigma$, which can be denoted as $\lambda \sim \sigma^4$. $\lambda$ is controlled by the proportion of r0 to r1 in resistor network, that is, the value of r0 adjusts the width of the filter function once r1 is permanent. We search a series of Gaussian functions with different $\sigma$ for every r0 in the resistor network to obtain one whose curve is the most closed to the network's. Figure 8 describes the deviation between an ideal Gaussian function and the resistor network, where we can see the relative error between them are almost less than 5%. We assign the number of the nodes in the resistor network as 45, and the proportion of r0 to r1 is 36, where the average error between ideal Gaussian curve and the resistor network is about 1.31%.

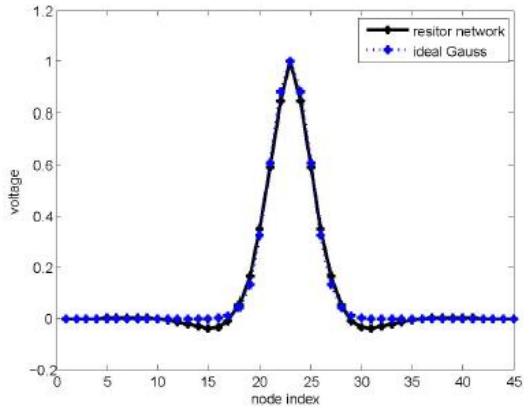

Fig. 8. Deviation between the filtered results of 1-D resistor network and ideal Gaussian function. There are 45 nodes in the 1D network, and the proportion of r0 and r1 is 36.

*B. Experiments on 2-D Gaussian Resistor Network*

In the case of one dimension, the active network can describe Gaussian function in a rather high precision as previous analysis. Because of the asymmetry in circle, the system function of 2-D resistor network appears high rate of descent. To analyze the deviation between 2-D Gaussian function and the system function of 2-D active resistor network, a signal pulse voltage is given to the network. The deviation between the filtered results of 2-D active resistor network and ideal Gaussian function are shown in figure 9. The x-coordinates represent the pixel-circle index ranging from the central voltage driver to outside. It is obvious those node-voltages closed to the center driver are higher than those far from it. The error of the first circle pixels surrounding the driver voltage is the highest one with a range from 12% to 16%, and the outer circles are almost less than 10%. Because of the impossibility to implement absolutely circle-symmetry circuit in 2-D network, the curve of the analog circuit descends more quickly than ideal Gaussian curve. However, considering the topology of the network, making more efforts to reduce the deviation is not a wise choice. The deviation can be taken as a bias, every input voltage signal will go through that, and then the bias can be counteracted in some degree.

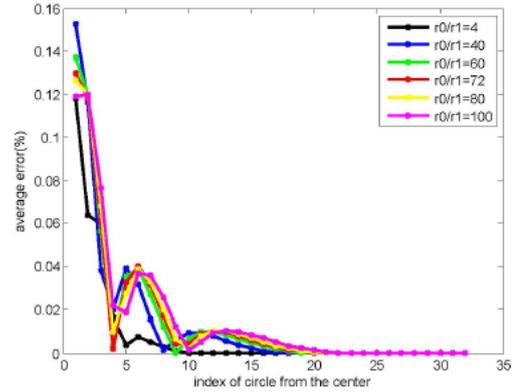

Fig. 9. Deviation between the filtered results of 2-D resistor network and ideal Gaussian function. The relative errors are different from each other with different r0, and the highest error between the network and ideal Gaussian is about 12% to 16%.

TABLE II
SETTING TIME OF THE ACTIVE RESISTOR FOR VARIOUS VALUES OF STRAY CAPACITANCE

| Value of stray capacitance(pf) | circuit setting time(ns) |
|---|---|
| 0.1 | 0.479 |
| 1 | 4.77 |
| 10 | 36.18 |
| 100 | 521.37 |

The active resistor consists of a number of resistors, including positive resistors and negative ones, and lots of voltages. Considering that there is no capacitive device in the circuit, the circuit setup time is very short. The principal factor to impact the setup time is the stray capacitances in MOSFETs, which are introduced by the active circuit part in the resistor network. Many previous works have demonstrated that the value of stray capacitance is relevant to the work frequency of the MOSFET. The work in [11] has shown the values of most stray capacitances stay on $fF-level(10^{-15}F)$. In this case, we use some common capacitances to simulate the stray capacitances. And we have verified the capacitance as several different values in table II.

In the analog Gaussian filter network, the positive resitors labeled as R1 in figure 6 are assigned as $250\,\Omega$, $R_2s$, controlling the filter width of the network, are variable in the analog Gaussian pyramid, and the range of the value is from 1k to 30k. Some simulations have been tried to test the setting time of the circuit. We can see the setting time grows quickly with the increasing of the stray capacitance. Considering the low work frequency of the Gaussian circuit system, the value of stray capacitance should be on $fF - level(10^{-15}F)$. As a result, the physical computing time is on nanosecond level or even less than it.

The power consumption of the circuit network is relative to the width of filter, which is illustrated as table III. With the increasing of $\lambda$, the power of each pixel decreases. As previously described, Gaussian pyramid is built by three octaves, each of which includes six different scale-filtered images. In our design, the values of $\lambda$ are assigned as {4, 20, 40, 80, 100, 120} respectively, and each $\lambda$ corresponds to an image in every octave. Totally, there are $86016(256 \times 256 + 128 \times 128 + 64 \times 64)$ pixels for every $\lambda$ in this Gaussian pyramid. We roughly estimate the setting time as 1ns, in this case, the power consumption of the pyramid is about 669.6 pico-Joule.

## V. Summary and Conclusions

Physical computing has great advantages on computing speed and power consumption when compared with conventional digital methods. Some algorithms are designed and optimized in the history, and some parallel techonologies, such as GPU and FPGA, are utilized to accelerate the processing in some specific applications. This paper presents an analog circuit which is designed to accelerate the SIFT algorithm, where an active resistor network is utilized to complete the Gaussian pyramid stage in SIFT. Voltages which are transformed from illumination by sensors correspond to each pixel, and the subsequent processing is directly at the source end. Different from analog computer, physical computing focuses on the natural characteristics of the devices and their topologies. In this way, a computing task is completed during the setting time of the circuit. By replacing the Gaussian pyramid stage with physical circuit, with no clock, SIFT has a great speed up, where just several nanoseconds or even less time need to set up the Gaussian pyramid, and more, the power consumption is just about 669:6pJ to the pyramid. Considering the relative low precision of the analog domain, this method is appropriate for some specific applications that need no high precision.